\documentclass[10pt,twocolumn,letterpaper]{article}

\usepackage{cvpr}              % To produce the CAMERA-READY version
\usepackage{multirow}
\usepackage{colortbl}
\usepackage{makecell}
\usepackage{arydshln}
\usepackage{bm}
\usepackage{marvosym}
\definecolor{cvprblue}{rgb}{0.21,0.49,0.74}
\usepackage[pagebackref,breaklinks,colorlinks,allcolors=cvprblue]{hyperref}
\usepackage{graphicx}
\def\paperID{8827} % *** Enter the Paper ID here
\def\confName{CVPR}
\def\confYear{2025}

\title{Bench-CoE: a Framework for Collaboration of Experts from Benchmark}

\author{Yuanshuai Wang$^{1,}$\thanks{denotes~ equal ~contributor;~\textrm{\Letter}~denotes corresponding author.} \quad Xingjian Zhang$^{1,\ast}$ \quad Jinkun Zhao$^{1,\ast}$  \quad Siwei Wen$^{1}$ \quad Peilin Feng$^{1}$  \quad Shuhao Liao$^{1}$\\
Lei Huang$^{1,2,~\textrm{\Letter}}$, \quad Wenjun Wu$^{1,2,~\textrm{\Letter}}$
\\
$^{1}$SKLCCSE, Institute of Artificial Intelligence,  Beihang University\\
$^{2}$Beijing Advanced Innovation Center for Future Blockchain and Privacy Computing,  Beihang University\\
\\
\normalsize\texttt{\{huangleiai, wwj09315\}@buaa.edu.cn}
}

\begin{document}
\maketitle
\begin{abstract}
Large Language Models (LLMs) are key technologies driving intelligent systems to handle multiple tasks. To meet the demands of various tasks, an increasing number of LLMs-driven experts with diverse capabilities have been developed, accompanied by corresponding benchmarks to evaluate their performance. This paper proposes the Bench-CoE framework, which enables Collaboration of Experts (CoE) by effectively leveraging benchmark evaluations to achieve optimal performance across various tasks. Bench-CoE includes a set of expert models, a router for assigning tasks to corresponding experts, and a benchmark dataset for training the router. Moreover, we formulate  Query-Level and Subject-Level approaches  based on our framework, and analyze the merits and drawbacks of these two approaches.  Finally, we conduct a series of  experiments with vary data distributions on both language and multimodal tasks to validate that our proposed Bench-CoE outperforms any single model in terms of overall performance. We hope this method serves as a baseline for further research in this area. The code is available at \url{https://github.com/ZhangXJ199/Bench-CoE}.

\end{abstract}
\section{Introduction}
\label{sec:intro}

Large Language Models (LLMs) are capable of performing various natural language processing (NLP) tasks, by using auto-regressive prediction conditioned by the task prompt \cite{gpt2, gpt3}.  LLMs' ability in describing and unifying tasks make them being the key components in current visual understanding tasks, which gives rise to the Large Multimodal Models (LMMs) \cite{llava, minigpt4}. While these models are able to perform all kinds of visual and language tasks, they may have different expertise and show significant diversity in performance for different tasks. We refer to these LLMs or LMMs models as experts in this paper.  One interesting question arises that how can we effectively identify and exploit the abilities of different experts. 
%These diversities of LLMs are likely from the diversity in training data, or the requirements while fine-tuning for specific domains. 

% A bunch of benchmark has been initially proposed to evaluate the performance of LLMs in certain tasks \cite{conll2003, snli, squad}. 
% Benchmarks are becoming more diverse, offering increasingly comprehensive evaluations of model capabilities, as the rapid progress. These include benchmarks such as MMLU\cite{mmlupro} for language tasks and MMMU \cite{mmmu} for multimodal tasks, both designed to assess models across multiple disciplines. Additionally, there are benchmarks for specific domains, such as GSM8K\cite{cobbe2021gsm8k} for mathematics and VCR\cite{zhang2024vcr} for visual reasoning. As a result, the performance display for LLMs is shifting from a single score to a more detailed ranking system that reflects models' strengths and weaknesses.

\begin{figure}[t]
  \centering
  \includegraphics[width=1.0 \columnwidth]{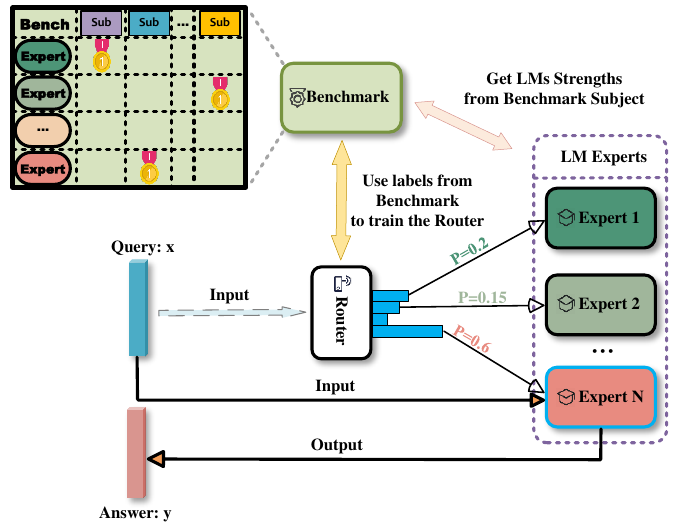}
  \caption{The framework of Bench-CoE. Our Bench-CoE Framework directly trains the router based on the benchmark, utilizing either subject-level or query-level labels for task assignment. This approach enables Bench-CoE to seamlessly integrate multiple expert models without incurring additional training costs, while simultaneously enhancing task performance.}
  \label{fig:2}
\end{figure}

 As the capabilities of experts gradually improve, benchmark tests have also become more complex and diverse\cite{conll2003, snli, squad}. For example, there are specific domain benchmarks such as GSM8K \cite{cobbe2021gsm8k} for assessing mathematical abilities and VCR \cite{zhang2024vcr} for evaluating visual reasoning. Additionally, there are benchmarks like MMLU \cite{mmlupro} for assessing language task multi-subjects reasoning abilities and MMMU \cite{mmmu} for evaluating multi-domain reasoning in multimodal tasks, both designed to evaluate expert models' capabilities across multiple subjects or domains. These reflect the evolution of expert model evaluation from single to diverse criteria, and they provide insights into the strengths and weaknesses of experts across different tasks.
 
We propose the Bench-CoE framework, which  enables expert collaboration (CoE) by effectively leveraging the strengths of different experts from benchmark evaluations. Bench-CoE includes a set of expert models, a router for assigning tasks to corresponding experts, and a benchmark dataset for training the router.
Based on this framework, we first formulate the  Query-Level Bench-CoE. The Query-Level approach is an abstraction of previous methods, such as\cite{routellm, routing2expert}. This approach requires evaluating the performance of different expert models for each query. Although it provides detailed information for training the router to select experts, it is expensive in terms of label creation and computational cost, and it struggles to generalize to tasks outside the data distribution. Is there a way to maintain generalization without additional cost? 

We analyzed that the key to the problem is obtaining labels for query assignment, and the performance of expert models across different subjects in benchmark tests actually serves as a type of label. Compared to the fine-grained Query-Level label, the more coarse-grained Subject-Level label is more likely to maintain the generalization of expert models. We further introduce Subject-Level Bench-CoE, which effectively exploits coarse-grained Subject-Level label from benchmark evaluations, as current benchmarks typically provide subject-level evaluation results \cite{mmlupro, mmmu}. During router training, each query in the benchmark dataset is used to train the router, which encourages the router to assign the query to the expert model that performs best in the subject of the query.

We evaluate the effectiveness and generalization of the query-level and subject-level routing learning mechanisms through a series of experiments. The experimental results show that both routing mechanisms improve the performance of the Bench-CoE model over using the best individual model. Specifically, the query-level router performs better on in-distribution data but is prone to overfitting on out-of-distribution data due to the fine-grained routing decisions. In contrast, the subject-level router demonstrates stronger generalization on out-of-distribution data, showcasing its better adaptability and robustness.

\begin{figure*}[t]
  \centering
  \includegraphics[width=2.1 \columnwidth]{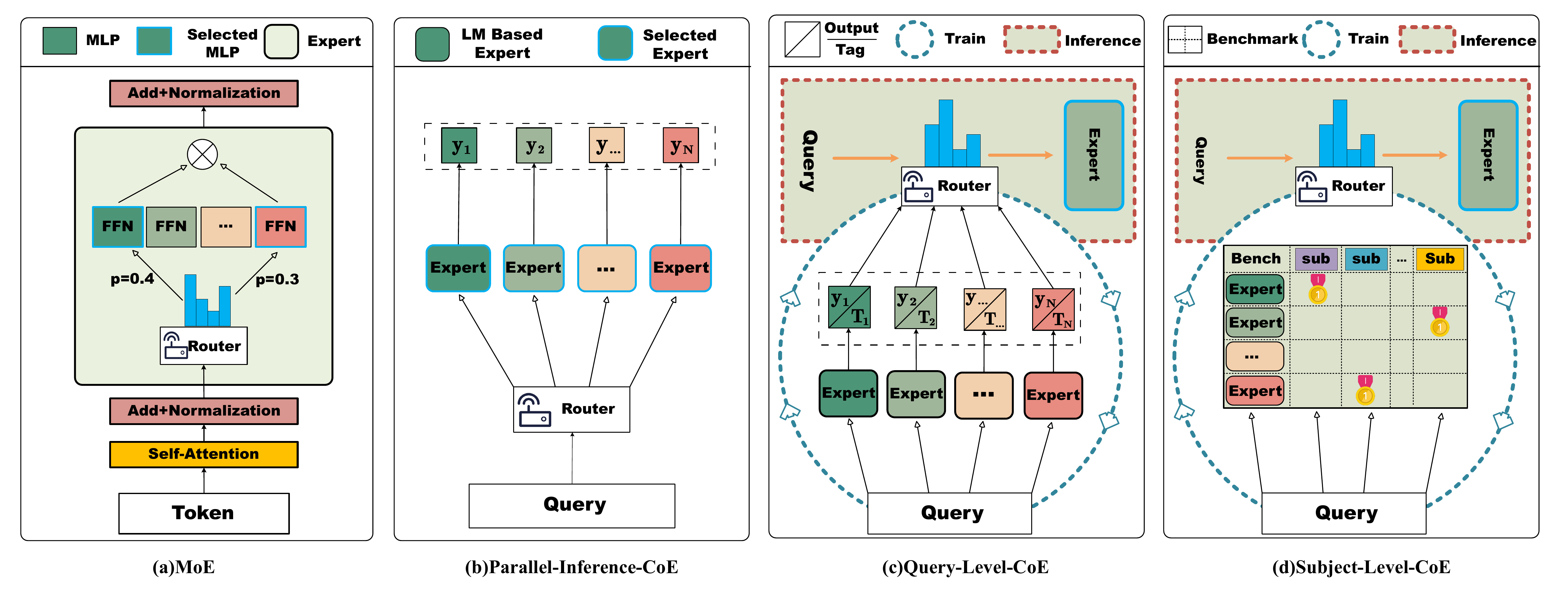}
   \caption{Comparison of routing methods in LLMs combination: (a) The MoE model utilizes multiple FFNs as expert modules during inference. (b) The Parallel-Inference-CoE model requires each query to pass through all experts during inference. (c) Although only the best expert is selected for inference, all expert models need to be tested during training to obtain labels. (d) Our Bench-CoE model only uses benchmark evaluation information to generate labels during router training, without extra costs, and uses only the best expert during inference.}
   
%    Comparison of routing methods in LLMs combination:
% (a) The MoE model utilizes multiple FFNs as expert modules during inference. (b) The Parallel-Inference-CoE model requires each query to pass through all experts during inference. (c) While only the best expert is selected for inference, all expert parameters are needed during training. (d) Our Bench-CoE model leverages only benchmark information during training without involving all experts and uses only the best expert during inference.
   \label{fig:l}
\end{figure*}

To summarize, our main contributions are as follows:

\begin{itemize}

\item We propose Bench-CoE, a simple and efficient pipeline for combining and routing LLM-driven experts, which achieves flexible and efficient task routing without relying on extensive labeled data and large-scale training.

\item We utilize the performance of each model on benchmarks to select the LLMs to be combined and construct subject-level and query-level datasets to support accurate routing for various specialized tasks.

\item Experiments demonstrate that the proposed method outperforms single models in multi-task scenarios, enhancing cross-domain multi-task processing performance with almost negligible inference cost.

\end{itemize}

\section{Related Work}
\label{sec:formatting}

% Efficiently utilizing and combining large language models (LLMs) in multi-task environments to address diverse query requirements and resource constraints has become a focus of recent research. Researchers have explored various routing and integration strategies, introducing methods to enhance the task-handling efficiency of LLMs. Existing methods can be broadly categorized as follows:

Recent research has focused on the efficient utilization and combination of LLMs in multi-task environments, aiming to meet diverse query requirements and manage resource constraints effectively. Researchers have investigated several routing and integration strategies, developing methods that improve the task-handling efficiency of LLMs. These existing methods can be broadly categorized as follows:

\subsection{Mixture of Experts} 
The Mixture of Experts (MoE) framework leverages multiple subnetworks as experts (e.g. FNN) with sparse activation, activating only a subset of experts per input to reduce computational costs while maintaining high performance. A milestone in MoE was achieved with Google's Switch Transformer \cite{SwitchTransformers}, which introduced a simplified routing mechanism to activate a few experts based on input features. Recent work has emphasized modularity to enhance MoE's adaptability. For example, RankMean \cite{rankmean} uses module-level importance scoring to efficiently merge fine-tuned models without access to training data. Similarly, routing mechanisms have gained prominence, with methods like PolyRouter \cite{polyrouter} extending the MoE paradigm to multi-LLM querying via predictive routing, and HDMoLE \cite{hdmole} employing hierarchical routing and dynamic thresholds for multi-domain fine-tuning. Overall, MoE models are centered on expert sub-models, aiming to integrate the specialized capabilities of these experts by combining their parameters and submodules. However, certain parameter-sharing approaches in MoE models often lack sufficient interpretability regarding the specific roles and contributions of individual expert sub-models. This lack of transparency poses challenges in understanding the decision-making process and the functionality of each expert, which may limit the model's trustworthiness and applicability.

\subsection{Parallel Inference CoE} 
Parallel Inference CoE aims to reduce inference costs by balancing resource utilization across models. FrugalGPT \cite{frugalgpt} uses a cascading strategy where tasks are processed sequentially across LLMs, dynamically adjusting model usage to optimize costs. LLM-Blender \cite{llm_blender} enhances responses by combining multiple LLMs through pairwise ranking and fusion, though its scalability is constrained by its dependence on high-quality data. Similarly, Hybrid LLM \cite{hybridllm} routes simpler tasks to smaller models while reserving larger models for complex queries, but it often relies on extensive labeled data for effective routing decisions. GraphRouter \cite{graphrouter} introduces a graph-based framework for routing, treating task and model selection as an edge prediction problem, though periodic updates and additional training data are required for new models. Alternatively, Eagle \cite{eagle} offers a training-free routing mechanism that dynamically selects models using global and local ranking, excelling in real-time scenarios but facing limitations with highly specialized tasks.
Parallel inference models, while capable of providing insights into the overall capabilities of different expert sub-models, require the same input to be dispatched to each expert sub-model for independent inference during execution. This approach results in significant computational resource wastage, substantially reducing inference efficiency. The inefficiency becomes particularly pronounced in scenarios with high computational demands, highlighting the need for improved resource utilization strategies.
% These models often come with significant computational overhead due to the need for multiple model activations and complex decision-making processes during each inference. In contrast, our Bench-CoE mitigates this issue by selecting only one LLM for each inference task, eliminating the need for additional computational overhead while maintaining competitive performance.

% Evaluation-Sample-Based Training
% \textbf{Evaluation-Sample-Based Training} To optimize routing and reduce inference costs, ZOOTER\cite{routing2expert} utilizes reward-guided strategies by extracting reward information from training queries to direct routing decisions, effectively selecting the most suitable expert for each query and minimizing computational demands. However, this approach relies on reward models that may require fine-tuning for specific applications. Similarly, RouteLLM\cite{routellm} dynamically routes queries based on task difficulty, assigning simpler tasks to lighter models and more complex tasks to stronger models, balancing response quality and cost-effectiveness. While effective, RouteLLM often depends on large labeled datasets and preference data, adding complexity and limiting its use in data-scarce environments.

\subsection{Query-Level CoE} 
At the query level, recent methods optimize routing by tailoring expert selection for each input. ZOOTER \cite{routing2expert} employs reward-guided strategies, extracting reward signals from training queries to direct routing decisions and minimize computational overhead. However, its reliance on fine-tuned reward models can constrain its applicability. Similarly, RouteLLM \cite{routellm} dynamically assigns simpler tasks to lighter models and reserves stronger models for more complex queries, balancing cost and quality. However, its dependence on large labeled datasets and preference data limits its utility in data-scarce environments. Although query-level mixture-of-experts (CoE) models enable the most fine-grained utilization of expert sub-model capabilities, they require extensive datasets with prior annotations. Each expert sub-model must undergo inference testing on these large-scale datasets to evaluate its performance on specific data. However, the resulting capability assessments are highly sensitive to variations in data distribution~\cite{routellm}, which can adversely impact the model's generalization performance and limit its applicability across diverse scenarios.

In contrast, our subject-level Bench-CoE addresses this limitation by employing a specific router, which utilizes models specializing in specific subjects as labels for handling corresponding tasks. These labels can be directly obtained from the benchmark evaluation leaderboard, making the implementation process simple, flexible, and more generalized. This design not only ensures that model selection aligns with well-defined criteria but also facilitates an intuitive understanding of the routing process and each model's contribution to the overall task, achieving clear interpretability.

\section{Method}

% Methodology

In this section, we provide a detailed description of the Bench-CoE. We first formulate a comprehensive framework. Then, under this framework, we formulate two approaches for training routing using BenchMark: the query-level approach and the subjective-level approach. The query-level approach is an abstraction of some previous methods. However, this method requires instance-level testing to define the data labels for training the routing, which makes it difficult to generalize. To address the generalization issue, we further propose a new approach, the subjective-level approach. We will explain these in detail next.

% First, we discuss the query-level approach, which is similar to most current methods. However, this approach requires testing individual large models to generate labels for training the router. To address this, we propose an alternative method that eliminates the need for any testing. Instead, we directly obtain the subject-level labels from the performance of each subject on the benchmark, and use these to train a router.

% 3.1. Problem Definition and Notation
\subsection{Bench-CoE Definition and Notation}
\noindent\textbf{Bench-CoE} is an approach of expert collaboration. It enhances the performance of task processing through the router, which can be described as:

\[
o = f(x, \{ M_l \}_{l=1}^L, \text{R}(\theta)),
\]
where \( x \) is the input data, \( o \) represents the final output result. \( \{ M_l \}_{l=1}^L \) represents a set of expert models, each expert model \( M_l \) may focus on processing specific sub-tasks. \( \theta \) refers to the parameters of the router  R , which regulates the collaboration of the experts. The Bench-CoE selects the most suitable expert from a group of multiple experts to process the input, with the goal of achieving overall performance superior to that of a single expert model.

\noindent\textbf{Benchmark and Subjects} A set containing \(V\) benchmark datasets is defined as \(\{ \mathcal{D}_1, \mathcal{D}_2, \dots, \mathcal{D}_V \}\). Without loss of generality, a benchmark dataset \( \mathcal{D}_K\) may contain \( B \) subjects, defined as follows:

\begin{equation}
\mathcal{D}_K = \{ \mathcal{S}^{1}, \mathcal{S}^{2}, \dots, \mathcal{S}^{B} \}.
\label{eq:1}
\end{equation}
Each subject \( \mathcal{S}^b \) corresponds to a set as shown below:

\begin{equation}
\mathcal{S}^b = \{ (x_{i}^b, t_{i}^b) \}_{i=1}^{|b|},
\label{eq:1}
\end{equation}
where
  \( x_{i}^b \) represents the \( i \)-th input in subject \( \mathcal{S}^b \), \( t_{i}^b \) is the corresponding standard answer, \( |b| \) is the number of samples in subject \( \mathcal{S}^b \).

\noindent\textbf{Models Set and Performance} A set containing \( L \) LLMs is defined as follows:

\begin{equation}
\mathcal{M} = \{ M_1, M_2, \dots, M_L \}.
\label{eq:0}
\end{equation}
For an input query\( x_{i}^b \) , the output of  \( M_l \) is shown as follows:

% For each model \( M_n \in \mathcal{M} \), generate the output for each input \( x_{ki}^B \) as follows:
\begin{equation}
o_{l,i}^b = M_l(x_{i}^b).
\label{eq:2}
\end{equation}
 For each benchmark, there is a corresponding evaluation metric function \( P_K \) to assess the performance of \( M_l \) on the samples \( (x_{i}^b, t_{i}^b) \), defined as follows:
% Using an appropriate evaluation metric function \( P_B \), we compute the performance score of model \( M_n \) on sample \( (x_{ki}^B, t_{ki}^B) \):
\begin{equation}
p_{l,i}^{K,b} = P_K(o_{l,i}^b, t_{i}^b).
\label{eq:3}
\end{equation}

\noindent\textbf{Router Definition}
Let \( R_\theta^L(x) \) be a routing function. It is parameterized by \( \theta \) and outputs a probability distribution over \( L \)  models:

\begin{equation}
R_\theta^L(x) = \text{R}(x, \theta, L),
\label{eq:10}
\end{equation}
where \( R_\theta^L(x) \) denotes the probabilities of \( L \) models to process the input \( x \). For text inputs,   Router R can be a BERT classifier\cite{bert}; for multi-modal inputs,  Router R can be a visual language model.

% 3.2. Model Performance Evaluation
% \subsection{Model Performance Evaluation}

% 3.3. Deriving Expertise Labels
% There are different ways to implement CoE. The current approach achieves CoE by processing queries to obtain a Large Language Model (LLM) specialized for that query, and then training a router using these as labels. Therefore, we refer to this method as Query-Level Bench-CoE. We will now provide a detailed explanation of this type.
Based on the framework formulated above, we refine two approaches that are query-level approach and subjective-level approach. 

\subsection{Query-Level Bench-CoE}
\label{subsec: 3.3}

\noindent\textbf{Query Label}
The performance of each model on each query \( x_{i}^b \) can be evaluated through test evaluation. The id of the model with the best performance is designated as the query-level label for that query, as defined below.

\begin{equation}
y_{i}^{K,b} = \arg\max_{l} p_{l,i}^{K,b}.
\label{eq:5}
\end{equation}
If multiple models achieve optimal performance on the same query, we select the model with the best overall performance on the benchmark.

% 3.4. Constructing the Router Training Dataset
\noindent\textbf{Router Dataset }
% Using the derived expertise labels, construct the dataset to train the router.
Once the label for each query are obtained, the benchmark dataset can be used to construct the query-level dataset \(\mathcal{D}_{query}\), represented as:

\begin{equation}
\mathcal{D}_{\text{query}} = \left\{ \left( x_{i}^{K,b}, y_{i}^{K,b} \right) \mid b = 1, \dots, B; \ i = 1, \dots, |b| \right\}.
\label{eq:8}
\end{equation}

\noindent\textbf{Router Train} 
The dataset can be used to train the router, and the expression for the query-level router loss function  \(\mathcal{L}_{\text{query}}(\theta)  \) is as follows:

\begin{equation}
\mathcal{L}_{\text{query}}(\theta) =  \sum_{b=1}^{B} \sum_{i=1}^{|b|} \ell(R_\theta^L(x_{i}^{K,b}), y_{i}^{K,b}).
\label{eq:6}
\end{equation}

\noindent\textbf{Model Set} 
Theoretically, to achieve the best combination results, the performance of all large models should be tested on each individual query. However, this  is not practical. Since the models selected based on subject specialization are already the best within their respective fields, we can directly choose them for combination. Experimental results show that this simplified approach is effective.

\noindent\textbf{Inference} 
For each new input \(x\), the router identifies the optimal model by predicting the model best suited to handle the input. The query-level model routing process is as follows:

\begin{equation}
\hat{n}_{\text{query}} = \arg\max_{l} R_\theta^L(x).
\label{eq:14}
\end{equation}
The Bench-CoE output using the query-level approach is shown as follows:

\begin{equation}
o = f(x,  M_{\hat{n}_{\text{query}}}, \text{R}(\theta)).
\label{eq:16}
\end{equation}

The Query-Level Bench-CoE method selects an LLM specialized in processing a given query, achieving good results on data within the same distribution. However, it requires testing the performance of numerous LLMs on a large dataset just like other query-level models\cite{routing2expert,routellm}, which can be challenging, and this approach struggles to maintain generalization on out-of-distribution data. We wondered if there could be a way to achieve CoE without the need for extensive testing. Upon analyzing this, we realized that the key to achieving this is to obtain an LLM specialized in handling specific queries. When we examine the performance of various LLMs on benchmark subjects, this can actually serve as a type of label—a label at the subject level. This insight leads us to the next approach we propose: Subject-Level Bench-CoE.

\subsection{Subject-Level Bench-CoE}
\noindent\textbf{Subject Label}
For each input \( x_{i}^b \) in the subject \( \mathcal{S}^b \),  we can directly obtain the subject-level label \( y_{i}^{K,b} \) for the query from the benchmark. Since the benchmark has already provided the following computational results when the leaderboard were released, our method does not require any additional manual data labeling.

\begin{equation}
y_{i}^{K,b} = \arg\max_{l} \frac{1}{|b|} \sum_{i=1}^{|b|} p_{l,i}^{K,b}.
\label{eq:4}
\end{equation}

\noindent\textbf{Router Dataset}
Once the subject-level labels \(D_{subject}\) for the queries are determined, the subject-level routing dataset can be obtained, as shown below: 
\begin{equation}
\mathcal{D}_{\text{subject}} = \left\{ \left( x_{i}^{K,b}, y_{i}^{K,b} \right) \mid b = 1, \dots, B; \ i = 1, \dots, |b| \right\}.
\label{eq:7}
\end{equation}
This dataset ensures that queries under each subject share the same label, making the router route query based on subject-specific knowledge.

% Selection of Models for the Composite System
\noindent\textbf{Model Set}
From the subject-level routing dataset, we select the appropriate models \(M_l\) to incorporate into our collaborative system, which can be represented as follows:
\begin{equation}
\mathcal{M}_{\text{set}} = \left\{  M_{l} \mid  l = y_{i}^{K,b} ,  y_{i}^{K,b}\in \mathcal{D}_{\text{subject}} \right\}.
\label{eq:9}
\end{equation}

% \noindent\textbf{Loss Function} 
% Unlike the query-level loss function, the subject-level loss function is as follows:

% \begin{equation}
% \mathcal{L}_{\text{subject}}(\theta) = \sum_{k=1}^{U} \sum_{i=1}^{n_k^B} \ell(R_\theta(x_{n,ki}^B), y_{ki}^B).
% \label{eq:6}
% \end{equation}
Once the dataset and model set are determined, the other processes remain the same as before, such as Router Train and Inference.

% , while the routing process of the subject-level model is as follows:
% \begin{equation}
% \hat{n}_{\text{subject}} = \arg\max_{u} R_\theta^U(x).
% \label{eq:14}
% \end{equation}

% 3.8. Experimental Setup
\subsection{Evaluation Scenarios}

To validate our approach, we designed three types of evaluation scenarios for both language and multimodal tasks, as follows.

% We conducted extensive experiments to validate our approach. For both language and multimodal tasks, we designed three types of experiments as follows.

%\noindent{\textbf{Scenario 1: Naive Evaluation}}
% naive test

% Since there is only one subset in MMLU Pro and MMMU, we used test dataset as the model benchmark in the training of router.
\paragraph{Naive Evaluation Scenario.}
Given that the MMLU Pro and MMMU datasets contain only one subset, we utilized the test dataset as the benchmark for router training.
% Router Training Data: Construct the router training dataset from \( \mathcal{D}_{\text{test}} \)  as described in \Cref{sec:experiments}. 
% Model Evaluation: Evaluate both composite and individual model performance on \( \mathcal{D}_{\text{test}} \). As the benchmark size gradually increases to the point where it covers various subejects, the significance of our proposed method becomes more apparent in this scenario. By simply leveraging benchmark information, we can combine models to achieve superior performance.
The router training dataset was constructed from \( \mathcal{D}_{\text{test}} \) as described in \Cref{sec:experiments}. 
Both composite and individual model performances were evaluated on \( \mathcal{D}_{\text{test}} \). As the benchmark size increased to encompass various subdomains, the advantages of this scenario became increasingly evident. By leveraging benchmark information, models were combined to achieve superior performance.

%\noindent{\textbf{Scenario 2: In-distribution Evaluation}}
% in-destributaion test

\paragraph{In-distribution Evaluation Scenario.} 
We utilized the training and validation subsets of the benchmark dataset respectively in the router training and testing phases.
The router training dataset was constructed from \( \mathcal{D}_{\text{train}} \) as described in \Cref{sec:experiments}. 
 Both composite and individual model performances were evaluated on \( \mathcal{D}_{\text{val}} \). This approach tested the performance of the combined model under the same data distribution but with different data splits, enhancing its generalization compared to the first scenario.

\paragraph{Out-of-distribution Evaluation Scenario.}
The router training dataset was constructed from \( \mathcal{D}_1 \) as described in \Cref{sec:experiments}. 
Both composite and individual model performances were assessed on \( \mathcal{D}_2 \). This approach tested the performance of the combined model under varying data distributions, further evaluating its generalization.

These experiments validated our approach across different tasks and data distributions. The results demonstrated that our method could enhance composite model performance without the need for extensive training or complex labels, consistently outperforming individual models across multiple benchmarks.

% Router Training Data: Construct the router training dataset from \( \mathcal{D}_1 \)  as described in \Cref{sec:experiments}. Model Evaluation: Evaluate both composite and individual model performance on \( \mathcal{D}_2 \) . This approach validates the performance of the combined model under different data distributions, further enhancing generalization.

% Through these experiments, we validated our approach across different tasks and data distributions. Results show that our method can improve composite model performance without requiring extensive training or complex labels, consistently outperforming individual models on multiple benchmarks.

\section{Experiments}
\label{sec:experiments}

% We conducted extensive experiments on both language and multimodal tasks to validate the effectiveness of our proposed method. The experiments aimed to evaluate the performance of our composite model (CoE) compared to individual LLMs under various settings, demonstrating the versatility and robustness of our approach.
We conducted extensive experiments on both language and multimodal tasks to validate the effectiveness of our proposed method. The experiments were designed to assess the performance of our Bench-CoE model against individual LLMs across various settings, demonstrating the versatility and robustness of our approach.

\Cref{tab:datasets_analysis} summarizes the characteristics of the datasets used in our experiments. It details whether each dataset contains multiple subsets, which is crucial for understanding the diversity and complexity of the data, as well as whether the datasets are annotated with subject labels, indicating their suitability for supervised learning tasks.
\begin{table}[h]
\centering
\begin{tabular}{lcccc}
\hline
Dataset & Train & Val & Test & Has Subject \\
\hline
MMLU Pro\cite{mmlupro} & No & Yes & Yes & Yes \\
%Hellaswag & Yes & Yes & Yes & No \\
Winogrande\cite{winogrande} & Yes & Yes & Yes & No \\
Big Bench Hard\cite{bbh} & No & No & Yes & No \\
MMMU\cite{mmmu} & No & Yes & Yes & Yes  \\
MMstar\cite{mmstar} & No & Yes & No & Yes  \\
\hline
\end{tabular}
\caption{Analysis of Datasets Characteristics.}
\label{tab:datasets_analysis}
\end{table}
These datasets were selected to provide a comprehensive test bed that challenges the capabilities of our models under both homogeneous and heterogeneous conditions. This varied dataset setup allows us to rigorously evaluate the adaptability of the composite model in scenarios ranging from closely related data subsets to entirely distinct data distributions. To assess the performance of our method in different scenarios, we performed three sets of experiments on language tasks and multimodal tasks.

\subsection{Naive Evaluation}

% In language task experiment, we trained the Bench-CoE and evaluated it on the same subset of the benchmark dataset MMLU-Pro\cite{mmlupro}. This setup tests the ability of the router to select the best model when both router training data and testing data are identical. The test set of MMLU-Pro, consisting of diverse language understanding tasks, was used. We compared four individual LLMs and our composite model (CoE) using both subject-level and query-level routers. The performance of each model is presented in \Cref{tab: Performance on MMLU-Pro test set}. The performance across subjects is shown in \Cref{fig:3}.

% In the language task experiment, Bench-CoE was trained and evaluated on MMLU-Pro\cite{mmlupro}. This approach tests the router's capability to select the most suitable model under conditions where the training and testing datasets are identical. We conducted a comparative analysis of Bench-CoE utilizing both subject-level and query-level routers and four individual LLMs that comprise it. The performance of each model is detailed in \Cref{tab: Performance on MMLU-Pro test set}, and the results across different subjects are illustrated in \Cref{fig:3}.

\begin{figure}[t]
  \centering
  \includegraphics[width=1.0 \columnwidth]{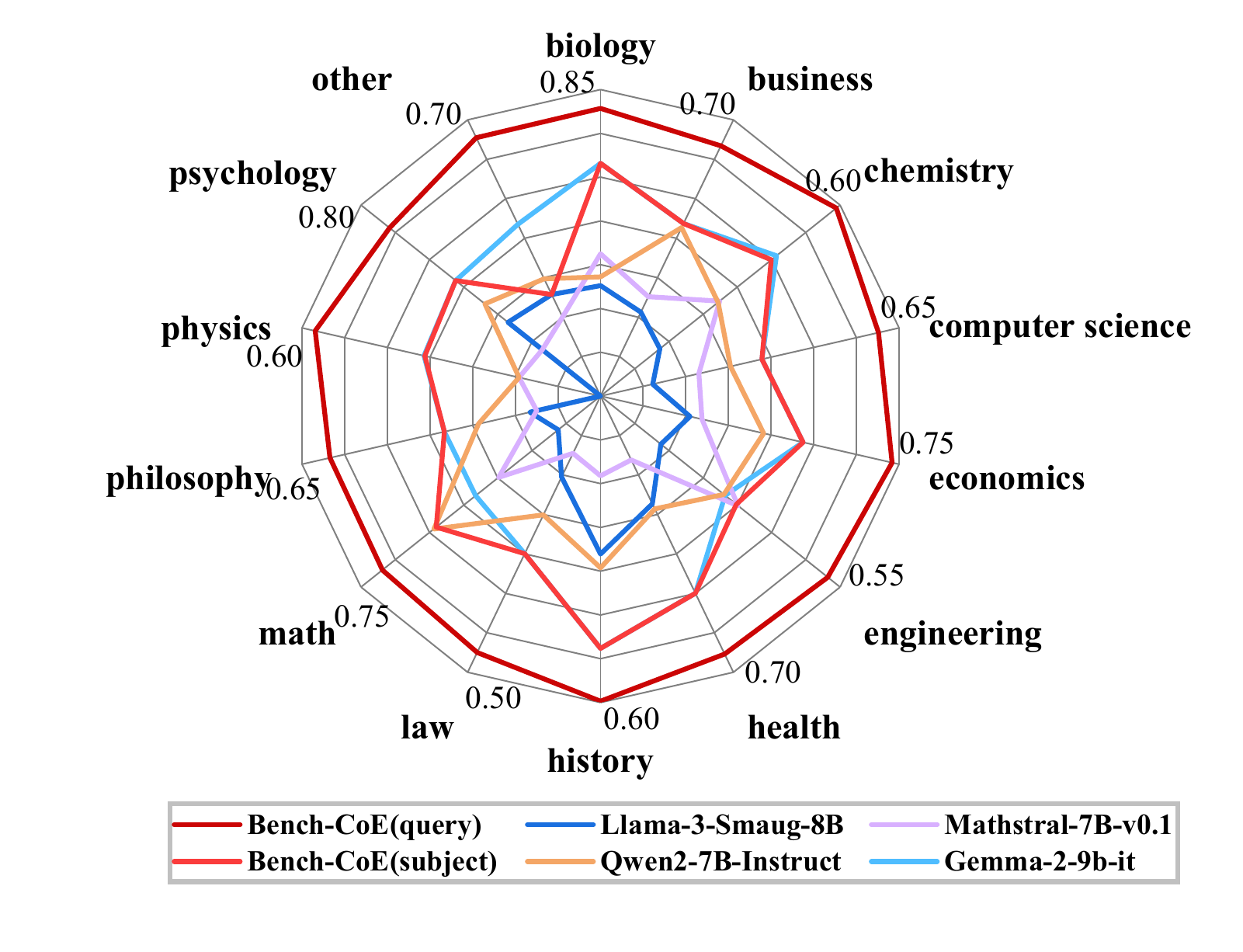}
  \caption{Performance Across Subjects on MMLU Pro. Bench-CoE (Query-Level) outperforms all other models comprehensively. Bench-CoE (Subject-Level) achieves performance comparable to the top MoE model, Gemma-2-9b-it, and outperforms it in certain subjects.}
  \label{fig:3}
\end{figure}

In the language task experiment, Bench-CoE was trained and evaluated on the same dataset MMLU-Pro\cite{mmlupro}. This setup tests the router's ability to select the most suitable model when both training and testing datasets are the same. We performed a comparative analysis between Bench-CoE, employing both subject-level and query-level routers, and these four individual LLMs that constitute it. Detailed performance metrics for each model are presented in \Cref{tab: Performance on MMLU-Pro test set}, while the results across various subjects are depicted in \Cref{fig:3}.

% ```latex
% Table 1. Performance on MMLU-Pro test set
\begin{table}[h]
\centering
\begin{tabular}{lcc}
\hline
Model & Accuracy & Increment\(\Delta{}\)  \\
\hline
Gemma-2-9b-it\cite{gemma} &  52.04\% &  0 \\
Llama-3-Smaug-8B\cite{llama3-smaug} & 38.10\% & - \\
Mathstral-7B-v0.1\cite{mistral} & 41.78\% & - \\
Qwen2-7B-Instruct\cite{qwen2} & 47.07\% & - \\
\textbf{Bench-CoE (Subject-Level)} & 52.24\% & 0.2\% \\
\textbf{Bench-CoE (Query-Level)} & \textbf{64.28\%} & \textbf{+12.24\%} \\
\hline
\end{tabular}
\caption{Performance on MMLU-Pro.}
\label{tab: Performance on MMLU-Pro test set}
\end{table}

% 111
The Bench-CoE with the query-level router achieved a performance of 64.28\%, significantly outperforming all individual models. The Bench-CoE with the subject-level router also surpassed individual models, though with a smaller margin. This demonstrates that Bench-CoE effectively leverages the strengths of different models when the data is consistent. The query-level router's finer-grained control allows it to select the best model for each specific input, leading to substantial performance gains, while the subject-level router improves performance by routing inputs to models generally better in specific subjects.

% The Bench-CoE with the query-level router achieved a performance of 0.6428, significantly surpassing all individual models. The version with the subject-level router also outperformed individual models, albeit with a narrower margin. These results demonstrate Bench-CoE’s capability to effectively harness the strengths of diverse models in a consistent data environment. The query-level router’s detailed control enables it to precisely select the optimal model for each input, leading to substantial performance improvements. Conversely, the subject-level router enhances performance by directing inputs to models that generally excel in specific subject areas.

\begin{figure}[t]
  \centering
  \includegraphics[width=1.0 \columnwidth]{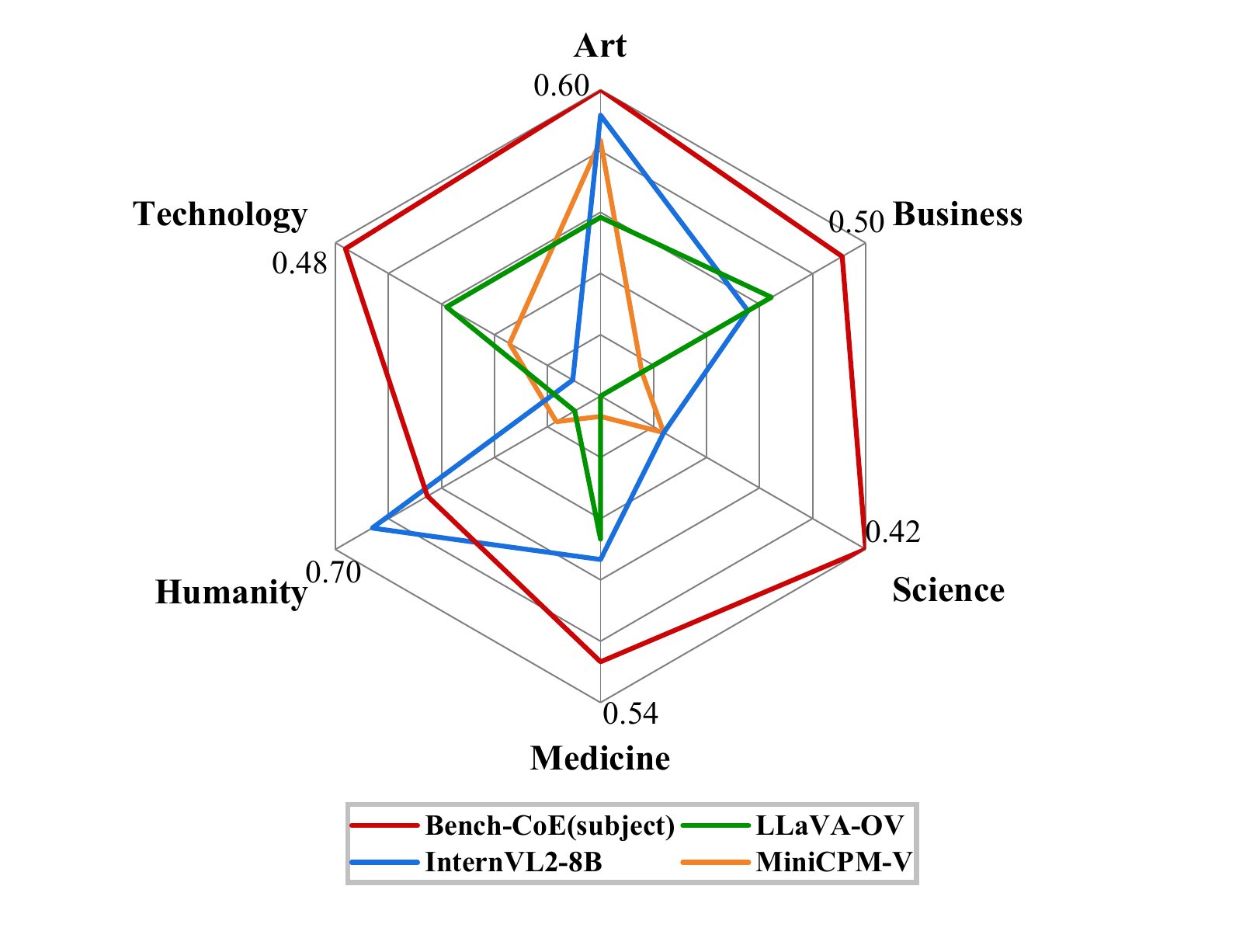}
  \caption{Performance Across Subjects on MMMU. Bench-CoE (Subject-Level) achieves significantly superior performance across almost all subjects.}
  \label{fig:4}
\end{figure}

In the multimodal task experiment, we trained the Bench-CoE and evaluated it on the same subset of the MMMU\cite{mmmu} dataset. We compared Bench-CoE with a subject-level router and three individual LLMs that constitute it. The results are presented in \Cref{tab: Performance on MMMU validation set}. The performance across subjects is shown in \Cref{fig:4}.

% Table 5. Performance on MMMU validation set
\begin{table}[h]
\centering
\begin{tabular}{lcc}
\hline
Model & Accuracy & Increment\(\Delta{}\) \\
\hline
MiniCPM-V-2.6 \cite{minicpm} & 45.22\% & - \\
InternVL2-8B \cite{internvl, internvl2} & 47.67\% & 0 \\
LLaVA-OV-7B \cite{LLaVA-OV} & 46.67\% & - \\
\textbf{Bench-CoE (Subject-Level)} & \textbf{51.78\%} & \textbf{+4.11\%} \\
\hline
\end{tabular}
\caption{Performance on MMMU.}
\label{tab: Performance on MMMU validation set}
\end{table}

% The CoE achieved a performance of 0.5178, significantly higher than all individual models. This demonstrates the effectiveness of our method in multimodal settings, effectively combining the strengths of different models.

The results demonstrate that Bench-CoE achieved a performance score of 51.78\%, which is significantly higher than that of all individual models. This result underscores the effectiveness of our approach in multimodal settings, as it successfully leverages the strengths of diverse models to enhance overall performance.

\begin{figure}[t]
  \centering
  \includegraphics[width=1.1 \columnwidth]{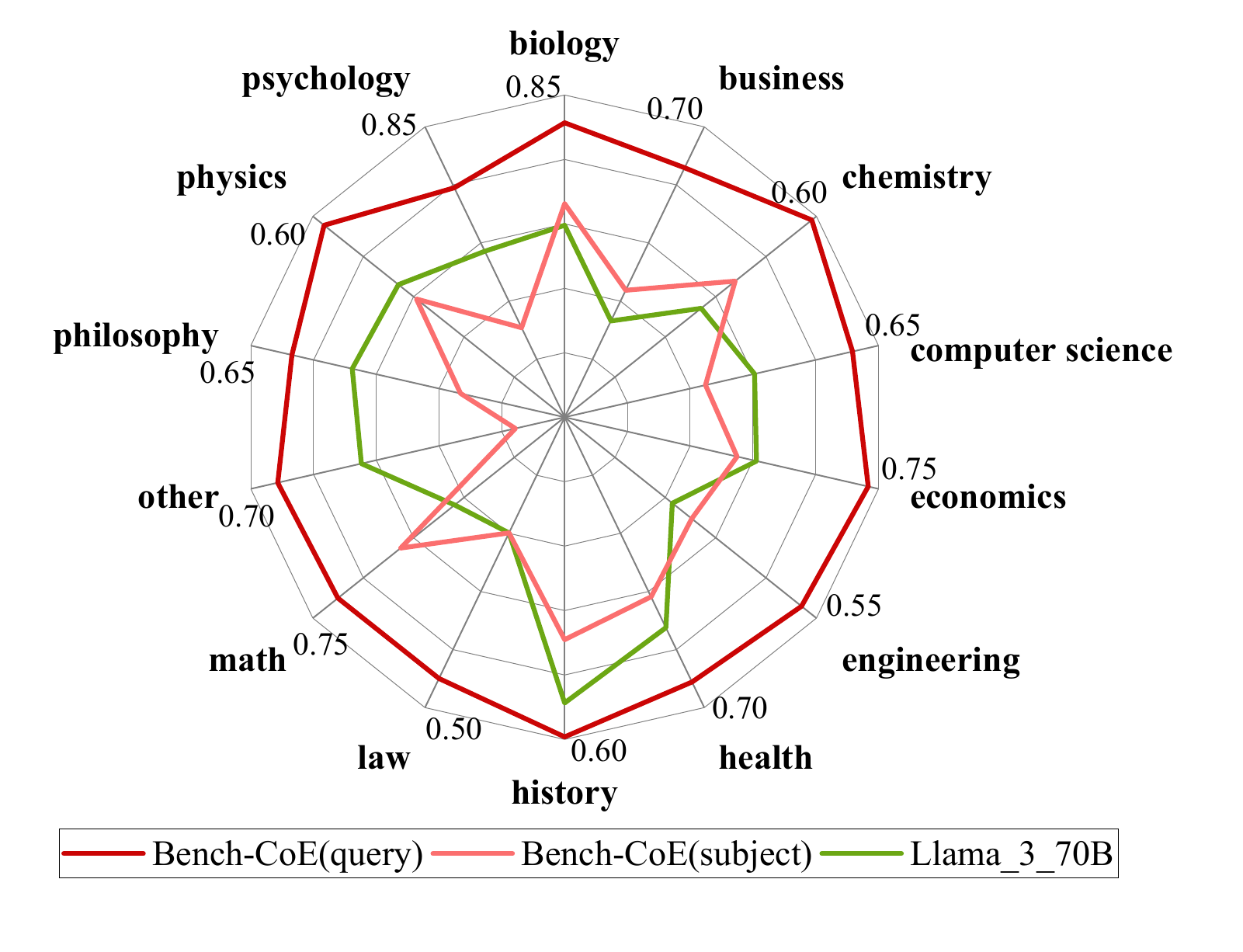}
  \caption{The performance of Llama-3-70B and Bench-CoE on each subject of the MMLU-Pro benchmark. Bench-CoE (Subject-Level) achieves performance comparable to Llama-3-70B, while Bench-CoE (Query-Level) surpasses Llama-3-70B.}
  \label{fig:Llama}
\end{figure}

\paragraph{Comparision to Larger LLMs.}
We compare our Bench-CoE by routing several small-scale LLMs to certain LLMs with larger parameters Llama-3-70B~\cite{llama3_70b}. We route four individual models as \Cref{tab: Performance on MMLU-Pro test set}  by using our Bench-CoE. The results are shown in Figure~\ref{fig:Llama}. We find that our Bench-CoE (the largest model used is only 9B) better than Llama-3-70B. 

\begin{figure}[t]
  \centering
  \includegraphics[width=1.1 \columnwidth]{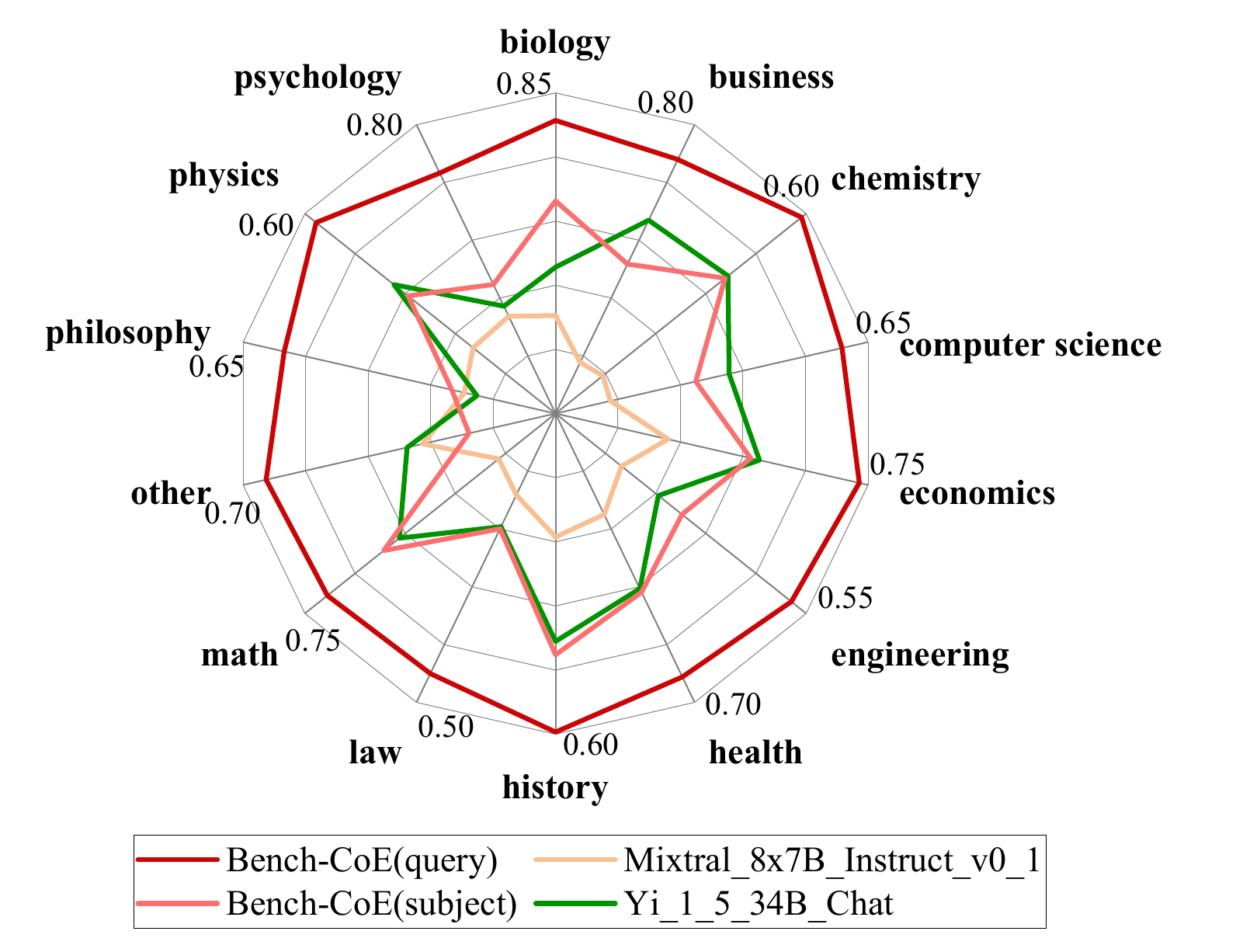}
  \caption{The performance comparison between Yi-1.5-34B-Chat, Mixtral-8x7B-Instruct-v0.1, and Bench-CoE. Bench-CoE (Query-Level) surpasses the other two models across all subjects.
  }
  \label{fig:Moe}
\end{figure}

We also compare our Bench-CoE to Mixtral-8x7B-Instruct-vO.1~\cite{mixtral8x7B} and Yi-1.5-34B-Chat~\cite{yi34b}, which uses MoE with larger parameters. The results are shown in Figure~\ref{fig:Moe}. Our Bench-CoE consistently obtains better performance than Mixtral-8x7B-Instruct-vO.1 and Yi-1.5-34B-Chat.

\subsection{In-distribution Evaluation}

%111
This experiment assesses the performance of the route when trained and tested on different splits of the same dataset. Specifically, the router was trained on the Winogrande training set and evaluated on the validation set. We conducted a comparative analysis of four individual LLMs and our CoE model using the query-level router. The results are presented in \Cref{tab: Performance on Winogrande validation set}.

% Table 2. Performance on Winogrande validation set
\begin{table}[h]
\centering
\begin{tabular}{lcc}
\hline
Model & Accuracy & Increment\(\Delta{}\) \\
\hline
Qwen2-7B-Instruct & 65.27\% & - \\
Gemma-2-9b-it & 66.14\% & 0 \\
Mathstral-7B-v0.1 & 55.95\% & - \\
Llama-3-Smaug-8B & 57.06\% & - \\
\textbf{Bench-CoE (Query-Level)} & \textbf{67.01\%} & \textbf{+0.87\%} \\
\hline
\end{tabular}
\caption{Performance on Winogrande.}
\label{tab: Performance on Winogrande validation set}
\end{table}

% According to the results, The Bench-CoE achieved the highest performance of 0.6701, outperforming all individual models. The improvement over the best individual model (Gemma-2-9b-it) is modest but consistent. Even when the router is trained and tested on different splits, it successfully generalizes and routes inputs to the most suitable models. The query-level router's ability to adapt to individual inputs contributes to the performance gain.

According to the results, Bench-CoE achieves the highest performance with an accuracy of 67.01\%, surpassing all individual models. The improvement over the best-performing individual model, Gemma-2-9b-it, is modest yet significant. Even when trained and tested on different splits, the Bench-CoE effectively generalizes and directs inputs to the most appropriate models. This adaptability of the query-level router contributes significantly to the performance gains.

%Additionally, we conducted a similar experiment on the HellaSwag\cite{hellaswag} benchmark. The results are as follows:

% Table 3. Performance on HellaSwag validation set
%\begin{table}[h]
%\centering
%\begin{tabular}{lcc}
%\hline
%Model & Accuracy & Increment\(\Delta{}\) \\
%\hline
%Qwen2-7B-Instruct & 0.7836 & - \\
%Gemma-2-9b-it & 0.8556 & 0 \\
%Mathstral-7B-v0.1 & 0.6240 & - \\
%Llama-3-Smaug-8B & 0.6304 & - \\
%\textbf{Bench-CoE (Query-Level)} & \textbf{0.8557} & \textbf{+0.0001} \\
%\hline
%\end{tabular}
%\caption{Performance on HellaSwag validation set.}
%\label{tab: Performance on HellaSwag validation set}
%\end{table}

% The CoE matches the performance of the best individual model (Gemma-2-9b-it), with a slight improvement. This suggests that our method can effectively combine models to achieve or surpass the best single-model performance.

% 111
% In the multimodal task experiment, we trained the router of Bench-CoE on the test set of MMMU and evaluated it on the validation set of MMMU. The consistent performance across different splits validates the effectiveness of our method. The results are shown in \Cref{tab: Performance on MMMU validation set (trained on test set)}.

In the multimodal task experiment, the Bench-CoE router was trained using the test set of MMMU and evaluated on the validation set of MMMU. The consistent performance across different dataset splits confirms the effectiveness of our approach. The detailed results are presented in \Cref{tab: Performance on MMMU validation set (trained on test set)}.

% Table 6. Performance on MMMU validation set (trained on test set)
\begin{table}[h]
\centering
\begin{tabular}{lcc}
\hline
Model & Accuracy & Increment\(\Delta{}\) \\
\hline
MiniCPM-V-2.6 & 45.22\% & - \\
InternVL2-8B & 47.67\% & 0 \\
LLaVA-OV-7B & 46.67\% & - \\
\textbf{Bench-CoE (Subject-Level)} & \textbf{50.78\%} & \textbf{+3.11\%} \\
\hline
\end{tabular}
\caption{Performance on MMMU.}
\label{tab: Performance on MMMU validation set (trained on test set)}
\end{table}

As shown in the table, Bench-CoE also outperforms individual models, achieving 50.78\%. The minimal performance drop compared to naive test indicates its effectiveness, even when the router is trained and tested on different splits.

\subsection{Out-of-distribution Evaluation}

To evaluate the generalization ability of Bench-CoE in language tasks, it was trained on the MMLU-Pro and tested on the validation set of Big-Bench-Hard\cite{bbh}. The evaluation included four individual LLMs that comprise Bench-CoE, along with the full Bench-CoE, using both subject-level and query-level routers. Performance results are detailed in \Cref{tab: Performance on Big-Bench-Hard validation set}.

% To test the generalization ability of our router, we trained it on one dataset and evaluated it on another. In language task experiments, the router was trained on the MMLU-Pro test set and evaluated on the Big-Bench-Hard\cite{bbh} validation set. We compared four individual LLMs and our Bench-CoE with both subject-level and query-level routers. The performance is summarized in \Cref{tab: Performance on Big-Bench-Hard validation set}.

% To evaluate the generalization capabilities of the Bench-CoE, it was trained on one dataset and tested on another. For language task experiments, the router was specifically trained using the MMLU-Pro test set and subsequently evaluated on the Big-Bench-Hard validation set, as cited in \cite{bbh}. We conducted comparisons between four individual Large Language Models (LLMs) that constitute the components of our Bench-CoE model and the Bench-CoE itself, employing both subject-level and query-level routers. The comparative performance is detailed in \Cref{tab: Performance on Big-Bench-Hard validation set}.

% Table 4. Performance on Big-Bench-Hard validation set
\begin{table}[h]
\centering
\begin{tabular}{lcc}
\hline
Model & Accuracy & Increment\(\Delta{}\) \\
\hline
Qwen2-7B-Instruct & 59.44\% & - \\
Gemma-2-9b-it & 65.10\% & - \\
Mathstral-7B-v0.1 & 66.35\% & 0 \\
Llama-3-Smaug-8B & 63.62\% & - \\
\textbf{Bench-CoE (Subject-Level)} & \textbf{69.91\%} & \textbf{+3.56\%} \\
\textbf{Bench-CoE (Query-Level)} & 67.07\% & +0.72\% \\
\hline
\end{tabular}
\caption{Performance on Big-Bench-Hard.}
\label{tab: Performance on Big-Bench-Hard validation set}
\end{table}

Based on the findings, Our Bench-CoE with the subject-level router achieved the highest performance of 69.91\%, outperforming all individual models. The query-level router also improved over individual models but performed slightly worse than the subject-level router in this cross-dataset scenario. The subject-level router generalizes better across different datasets because it relies on broader subject characteristics rather than specific input features, which may vary between datasets. The query-level router may overfit to the training dataset's specific input patterns, leading to slightly reduced performance on unseen data.

In multimodal task experiment, we trained the router on the MMMU validation set and evaluated it on the MMStar\cite{mmstar} dataset. The results are in \Cref{tab: Performance on MMStar}.

% Table 7. Performance on MMStar
\begin{table}[h]
\centering
\begin{tabular}{lcc}
\hline
Model & Accuracy & Increment\(\Delta{}\)\\
\hline
MiniCPM-V-2.6 & 54.33\% & - \\
InternVL2-8B & 59.22\% & 0 \\
LLaVA-OV-7B & 55.86\% & - \\
\textbf{Bench-CoE (Query-Level)} & 56.00\% & -3.22\% \\
\textbf{Bench-CoE (Subject-Level)} & \textbf{60.09\%} & \textbf{+0.87\%} \\
\hline
\end{tabular}
\caption{Performance on MMStar.}
\label{tab: Performance on MMStar}
\end{table}

As shown in the table, Bench-CoE with subject-level router achieved the highest performance of 60.09\%, surpassing the best individual model(InternVL2-8B). This confirms that our method generalizes well to different datasets in multimodal tasks, effectively leveraging the strengths of individual models. However, the performance of the Bench-CoE with query-level router has not surpassed that of the best model, which we attribute to the router relying solely on text input for classification, but many text inputs in the multimodal dataset are similar, and distinguishing query types requires image-based cues. Consequently, the generalization performance of Bench-CoE with the query-level router is suboptimal. Therefore, conducting experiments with a multimodal router will be one of our future directions.

\subsection{Overall Observations for Experiments}

In language task experiments, our method consistently outperforms individual models across different experimental setups. The choice between subject-level and query-level routers depends on the scenario: query-level routers excel when data distributions are similar, while subject-level routers generalize better across different datasets. In multimodal task experiments, our method effectively combines multimodal models to improve performance. The consistent performance gains across experiments validate the flexibility and robustness of our approach in handling different data modalities.

% \textbf{Discussion}
\section{Discussion}

\paragraph{Advantages of Bench-CoE.}
Our Bench-CoE model consistently outperforms individual models across a variety of tasks and datasets. This model is highly flexible, effectively addressing both language and multimodal tasks, and it does not require extensive training phases or hard-to-obtain labels. By utilizing benchmark performance to generate routing labels, Bench-CoE efficiently harnesses the strengths of different models without necessitating significant additional resource expenditures.

% Our composite model consistently outperforms individual models across various tasks and datasets. The method is flexible, working effectively for both language and multimodal tasks, and does not require extensive training or hard-to-obtain labels. By using benchmark performance to generate routing labels, we efficiently leverage the strengths of different models without significant additional resource requirements.

\paragraph{Reasons for Performance Gains.} Here, we provide three likely reason for why our Bench-CoE works well: 
\begin{itemize}
\item Leveraging Model Strengths. Different models excel in various subjects or on specific inputs. Bench-CoE capitalizes on these strengths by effectively routing each input to the most suitable model.

\item Effective Routing. The Bench-CoE router accurately predicts the best model for each input or subject, enhancing the overall system performance by ensuring efficient model allocation.

\item Generalization Ability. Particularly notable with subject-level routing, Bench-CoE demonstrates strong generalization to unseen data distributions, consistently maintaining robust performance across diverse datasets.
\end{itemize}

% \textbf{Comparison of Routers}

% - Query-Level Router: Provides fine-grained routing decisions, leading to higher performance when training and testing data are similar. However, it may overfit to the training data, reducing generalization across datasets.

% - Subject-Level Router: Offers better generalization across different datasets by focusing on broader subject characteristics. It may not capture the nuances of individual inputs as effectively as the query-level router but demonstrates robustness in cross-dataset scenarios.

% \textbf{Support for Our Proposed Method}

% The experimental results strongly support our proposed method, demonstrating that combining multiple LLMs via an effective routing mechanism leads to superior performance. The method is practical and efficient, as it does not require extensive additional resources beyond what is needed to evaluate individual models.

\paragraph{Limitations and Future Work.}
Although Bench-CoE demonstrates considerable potential, there remain opportunities for further enhancement:
\begin{itemize}
\item Router Complexity. Exploring more sophisticated routing models may further enhance performance, especially in scenarios where the query-level router overfits.

\item Scalability. Assessing the method's scalability with a larger number of models or on larger datasets would be valuable for real-world applications.

\item Dynamic Model Integration. Investigating how to dynamically add new models into the composite system without retraining the router from scratch could improve the method's adaptability.
\end{itemize}

% \textbf{Conclusion}
\section{Conclusion}

% Our experiments validate that the proposed Bench-CoE effectively combines multiple LLMs to achieve superior performance across various tasks and datasets. The method leverages the strengths of individual models without the need for extensive training or complex labels, providing a solid foundation for future research in model combination and routing strategies.
This paper proposed a simple framework for Collaboration of Experts (CoE) by effectively exploiting the evaluation from benchmarks. The formalized query-level and subject-level routing mechanism  effectively integrates multiple LLMs, delivering superior performance across diverse tasks and datasets. By harnessing the strengths of individual models without requiring extensive training or intricate labeling, Bench-CoE establishes a robust baseline for advancing research in model integration and routing strategies. 

\paragraph{Acknowledgment.}
This work was partially supported  by the National Key Research and Development Plan of China under Grant 2022ZD0116310, National Natural Science Foundation of China (Grant No. 62476016), the Fundamental Research Funds for the Central Universities.
{
    \small
    \bibliographystyle{ieeenat_fullname}
    \bibliography{main}
}
\clearpage
\appendix
\section{Models and Datasets}
\subsection{Language Task Models}
Qwen2-7B-Instruct is an instruction-focused language model developed by Qwen Technology. Designed to excel in various natural language understanding tasks, this model utilizes an optimized decoding strategy to enhance performance. With 7 billion parameters, it is well-suited for complex text comprehension and generation tasks, especially in Chinese contexts. Qwen2-7B-Instruct is particularly effective for instruction-responsive tasks such as content creation, information extraction, and dialogue systems.

Gemma-2-9b-it is a large language model developed by Gemma Technologies with 9 billion parameters, tailored for the information technology (IT) sector. Its training data encompasses a vast array of technical documents, programming guides, and texts from open-source projects. This model excels in understanding and generating highly specialized IT content, making it ideal for applications in technical support, documentation automation, and code parsing.

Mathstral-7B-v0.1 is a language model focused on solving mathematical problems, developed by the Mathstral team. With 7 billion parameters, its training includes extensive mathematical educational materials and real-world problem-solving examples. Mathstral-7B-v0.1 is designed to aid in mathematical education, automated problem-solving, and mathematical research, particularly effective for complex mathematical questions and theoretical discussions.

Llama-3-Smaug-8B is the latest large language model from the Llama team, featuring 8 billion parameters. It has been extensively pre-trained across multiple languages and domains to provide broad knowledge coverage and deep semantic understanding. Llama-3-Smaug-8B emphasizes performance in complex linguistic reasoning, long-form text generation, and multi-domain knowledge integration, suitable for advanced natural language processing tasks such as text summarization, language translation, and cross-domain knowledge-based question answering.

\subsection{MultiModal Task Models}
MiniCPM-V-2.6 is a multimodal language model developed to integrate visual processing with natural language understanding. With 2.6 billion parameters, this model is a compact version of the larger CPM series, designed to efficiently handle tasks that require the synthesis of textual and visual data. MiniCPM-V-2.6 excels in image captioning, visual question answering, and other applications where joint understanding of text and image is critical. Its training regimen includes diverse datasets from both textual and visual domains, ensuring robust performance across a variety of multimodal challenges.

InternVL2-8B is an 8 billion parameter model specifically designed for video-language tasks. Developed to bridge the gap between dynamic visual content and language, InternVL2-8B can analyze and generate descriptions for video data, making it highly suitable for applications such as automated video captioning, video content analysis, and interactive video-based learning systems. Its architecture allows for deep understanding of temporal video sequences in conjunction with textual descriptions, providing state-of-the-art results in video understanding tasks.

LLaVA-OV-7B, standing for Language and Vision Analysis - OmniVision, is a 7 billion parameter language model that specializes in comprehensive visual and textual interpretation. This model integrates advanced vision capabilities with natural language processing to perform tasks like detailed image analysis, multimodal translation, and cross-modal information retrieval. LLaVA-OV-7B is trained on a vast array of multimodal data sources, enabling it to effectively understand and generate content that requires the amalgamation of visual cues with textual data.

\subsection{Language Task Datasets}
MMLU-Pro is an extension of the original MMLU dataset, designed to evaluate language models on professional-level topics across a wide array of subjects. This dataset includes complex questions that require not only language understanding but also domain-specific knowledge, ranging from medicine and law to engineering and the arts. MMLU-Pro aims to test the depth and breadth of a model's understanding of advanced topics, making it a rigorous benchmark for language comprehension.

Winogrande is a large-scale dataset designed to improve the robustness and challenge of Winograd Schema Challenge-style tasks. It involves natural language inference tasks where the model must resolve ambiguity in sentences using common-sense reasoning. The dataset is particularly known for its difficulty and diversity, requiring models to utilize a deep understanding of context and world knowledge to make the correct inferences.

Big-Bench-Hard is a subset of the broader BIG-bench dataset specifically curated to challenge the capabilities of language models with particularly difficult tasks. This dataset includes a variety of language-based tasks such as analogical reasoning, complex problem-solving, and advanced comprehension challenges that go beyond the typical capabilities of standard language models, pushing the limits of what AI can understand and process in textual form.

\subsection{MultiMode Task Datasets}
MMMU is a comprehensive dataset designed for evaluating the performance of multimodal models across tasks that require simultaneous understanding of text, image, and sometimes audio content. This dataset includes challenges such as cross-modal retrieval, multimodal reasoning, and synchronizing visual content with textual descriptions. MMMU aims to simulate real-world scenarios where multiple types of data must be integrated and interpreted together.

MMStar is a multimodal dataset focused on the interplay between visual and textual data in entertainment and media contexts. It includes annotated images and videos from various media sources, coupled with descriptive texts and contextual information. The dataset is utilized for tasks such as multimedia content summarization, sentiment analysis, and thematic classification, testing a model's ability to navigate and interpret complex media-rich environments.

\section{Experiment Details}
\subsection{Language Experiment}
Due to the current limitations in large model evaluation techniques, there is a relative scarcity of benchmarks and datasets specifically tailored to academic disciplines. To the best of our knowledge, only the MMLU-Pro and Big-Bench-Hard datasets include manually annotated discipline-specific labels. This poses significant challenges to the experimental design of our Bench-CoE model. To thoroughly evaluate the performance of Bench-CoE, we conducted the following three types of tests:

During the naive test phase, we selected the MMLU-Pro dataset, which features well-defined discipline-specific labels, for training and evaluation of the BERT model. However, since the MMLU-Pro dataset only provides validation and test sets, we conducted both training and testing on the validation set. As the experiments and evaluations in this phase were performed on the same dataset, the results primarily serve to demonstrate the basic feasibility of our proposed approach. To further evaluate the effectiveness and generalizability of Bench-CoE, we designed more sophisticated experiments, including both in-distribution and out-of-distribution tests.

In the in-distribution test phase, we evaluated Bench-CoE using the Winogrande dataset, which provides a clear separation between training and test sets. Specifically, we trained the Bench-CoE model on the training set of Winogrande and tested it on the corresponding test set. However, since the Winogrande dataset lacks strong discipline-specific features (e.g., no manually annotated discipline labels), it was not possible to directly assess the model's capabilities through a discipline-wise leaderboard. As a result, we focused solely on evaluating the query-level performance of the Bench-CoE model.

In the out-of-distribution test phase, we selected datasets with strongly defined discipline-specific features: the MMLU-Pro dataset as the training set and the Big-Bench-Hard dataset as the test set. Specifically, we trained the Bench-CoE router on the MMLU-Pro dataset and evaluated it on the Big-Bench-Hard dataset. By testing across different datasets with distinct data distributions, and with both training and test sets exhibiting clear discipline-specific characteristics, this phase allowed us to thoroughly validate the generalization capability of the Bench-CoE model at both the query-level and subject-level.

\subsection{MultiModal Experiment}

MMMU and MMStar are currently among the most comprehensive multimodal benchmarks, encompassing tasks such as cross-modal retrieval and multimodal reasoning. To thoroughly evaluate the performance of Bench-CoE on multimodal tasks, we designed experiments in three phases: naive test, in-distribution test, and out-of-distribution test.

In the naive test phase, we used the MMMU dataset for both training and testing the Bench-CoE router. The subset of MMMU was utilized for both training and evaluation. This phase primarily aimed to verify the basic feasibility of Bench-CoE in task allocation for multimodal tasks. By leveraging query-level and subject-level routing strategies, Bench-CoE significantly outperformed individual models, demonstrating its effectiveness in task allocation. The query-level router provided finer-grained task assignments, while the subject-level router exhibited stronger overall robustness.

In the in-distribution test phase, the test set of the MMMU dataset was used for training, and the validation set was used for evaluation. This setup ensured a clear separation between training and testing data while maintaining consistency in data distribution. The Bench-CoE router effectively allocated tasks to the most suitable expert models based on the input, showcasing its strong adaptability for tasks within the same distribution.

In the out-of-distribution test phase, the Bench-CoE router was trained on the validation set of the MMMU dataset and tested on the MMStar dataset. The MMStar is a multimodal dataset focus on the interplay between visual and textual data in entertainment and media contexts, presenting challenges to the model’s generalization capabilities. The experiments demonstrated that the subject-level router remained effective in handling tasks with significant distributional differences, validating the adaptability and robustness of Bench-CoE. In contrast, the query-level router showed slightly reduced performance on new data distributions, likely due to overfitting.

These experimental results indicate that Bench-CoE effectively integrates the strengths of different models, achieving outstanding performance in both in-distribution and out-of-distribution tasks. This approach provides a solid foundation for further research on collaborative mechanisms in multimodal models.

\section{Scalability of Bench CoE}
In Bench-CoE, particularly in the subject-level Bench-CoE, we leverage the best-performing LLM for each domain as the routing target. By directing as many questions as possible within a given domain to the "best" LLM for inference, we enhance the overall accuracy of the model. However, with the rapid evolution of large language models, accompanied by the introduction of new datasets, novel models, and updated evaluation methods, the leaderboard rankings of LLMs change frequently. Under such circumstances, a fixed routing strategy in the CoE model cannot accommodate newly emerging models or adapt to shifting data distributions.

To address this limitation and improve the scalability of Bench-CoE, we designed a leaderboard-prior-based subject routing mechanism. Instead of directly routing inputs to a fixed best-performing model in a domain, our router first predicts the subject type of the given input. It then leverages the leaderboard-prior subject-to-model mapping to route the input to the latest and most suitable model for that domain. This approach significantly enhances the scalability of Bench-CoE, allowing it to flexibly adapt to rapidly evolving datasets and LLM advancements by dynamically adjusting the leaderboard and updating routing rules.

\section{Scenarios Unsuitable for CoE}
In our experiments with the Bench-CoE model, we selected a wide range of LLMs as candidate models and conducted extensive testing. Through these tests, we identified a common challenge in the CoE field: the issue of LLM capability diversity. Specifically, this problem arises when a candidate LLM lacks capability diversity on the given dataset—either significantly outperforming or underperforming all other candidate LLMs. Such cases negatively impact the overall performance of the CoE model, as the router is forced to route all queries either exclusively to or completely away from this model to achieve optimal results. This creates a significant challenge for training the router.

Looking ahead, we believe this issue can be mitigated with the development of dynamic routing strategies and adaptive candidate LLM selection mechanisms. These advancements will enable the CoE model to better handle capability imbalances among candidate LLMs, paving the way for more robust and flexible routing solutions.


\end{document}